\documentclass[11pt]{article}

\usepackage{ai_lab_2026}

\usepackage{times}
\usepackage{latexsym}

\usepackage[T1]{fontenc}

\usepackage[utf8]{inputenc}

\usepackage{microtype}

\usepackage{inconsolata}

\usepackage{graphicx}

\usepackage{amsmath}
\usepackage{amsfonts}
\usepackage{amssymb}

\usepackage{booktabs}
\usepackage{multirow}
\usepackage{xurl}  

\usepackage{xcolor}

\usepackage[
    colorlinks=true,
    linkcolor=black,   
    citecolor=blue,    
    urlcolor=blue      
]{hyperref}

\title{On-Policy Delta Distillation\\for Multilingual Math Reasoning}

\author{
    Byeongho Heo
    \quad
    Jaehui Hwang
    \quad
    Sangdoo Yun
    \quad
    Dongyoon Han
    \\ [2mm]
    NAVER AI Lab
}

\begin{document}
\maketitle
\begin{abstract}
On-Policy Distillation (OPD) is emerging as a promising alternative to reinforcement learning for LLM post-training, yet its effectiveness in multilingual settings remains underexplored. We study OPD and its advanced variant, On-Policy Delta Distillation (OPD$^2$), for mathematical reasoning in English, Korean, and Japanese. OPD$^2$ improves OPD by using the probability gap between a post-trained teacher and its base model as the learning signal. Experiments with Qwen3 show that OPD$^2$ consistently outperforms the original OPD, with particularly strong improvements in Korean and Japanese, and generally narrows the English--Korean performance gap. We further find that English-only OPD can also increase performance for Korean and Japanese, but often shifts the responses toward English, highlighting the importance of multilingual data to preserving target-language responses.
\end{abstract}

\section{Introduction}

The On-Policy Distillation (OPD) emerges as a promising alternative for Reinforcement Learning (RL) for LLM post-training. While RL evaluates the score of the whole generated responses with a single score that makes sequence-level supervision, OPD utilizes token-level supervision by inputting the student's rollout into the teacher model and getting token-level generation probabilities as feedback. This process addresses the limited representativeness of sequence-level supervision in RL-based post-training, improving data and computation efficiency. Based on this technique, recent studies have either adopted OPD for model post-training~\citep{yang2025qwen3,xiao2026mimo} or proposed variants of OPD~\citep{yang2026exopd,heo2026opd2}.

Recently, On-Policy Delta Distillation (OPD$^2$) \citep{heo2026opd2} improves the performance of OPD with the delta signal. Instead of the difference between teacher and student logit probabilities in the original OPD, OPD$^2$ utilizes the gap between the teacher and its base model as the learning signal for OPD. 
Despite this simple modification, OPD$^2$ substantially outperforms OPD on mathematical, scientific, and coding reasoning tasks across the Qwen3 model family.

Despite these advances, OPD remains underexplored in multilingual settings. 
Existing OPD studies have primarily focused on reasoning benchmarks in
English~\citep{jin2026eopd,yang2026exopd,song2026survey,heo2026opd2}.
Liu et al.~\cite{liu2026crosslingual} recently extended on-policy self-distillation to
African languages, demonstrating its potential for multilingual reasoning.
However, their study focuses on self-distillation, where the same model serves
as both the student and teacher, rather than OPD with a strong
teacher model. Moreover, the effectiveness of OPD for East Asian
languages, including Korean and Japanese, remains largely unexplored.

In this paper, we extend the study of OPD$^2$ to multilingual mathematical reasoning in Korean and Japanese. We empirically investigate whether OPD and OPD$^2$ can further improve the multilingual reasoning capabilities of strong models.
Specifically, we address the following research questions:

\begin{enumerate}
\item Does OPD$^2$ still outperform the original OPD in multilingual settings?
\item Does OPD$^2$ post-training help narrow the language gap in reasoning performance?
\item How does EN-only OPD affect multilingual reasoning and target-language generation?
\end{enumerate}

Our experimental results show that OPD$^2$ consistently outperforms the original OPD in multilingual settings. OPD$^2$ post-training also tends to reduce the performance gap between English and non-English languages, although this effect is not consistent across all models and settings. Interestingly, English-only OPD can still improve performance on non-English benchmarks. However, such gains do not necessarily indicate improved reasoning ability in the target language, as the generated reasoning process often partially relies on English even when the input problem is presented in a non-English language. These findings provide a broader understanding of multilingual OPD and offer a useful foundation for future research on extending OPD to languages beyond English.

\section{On-Policy Distillation}

We briefly explain the On-Policy Distillation (OPD) mechanism with formulations. 
OPD is basically used to replace the post-training process conducted by RL. 
Thus, it also shares the on-policy nature with RL post-training, such as GRPO~\citep{shao2024deepseekmath}.
Given a question $x$, the policy model $\pi_\theta$ (i.e., the student) first generates a response $y$ from its current policy.
The generated question–response pair is then fed into the teacher model $\pi^*$ to obtain the teacher’s token-level probability over the student-generated trajectory.
The student is trained to reduce the discrepancy between its output distribution and that of the teacher.
The OPD objective can be written as
\begin{equation}
\mathbb{E}_{y \sim \pi_{\theta}(\cdot \mid x)}
\left[
D_{\mathrm{KL}}
\left(
\pi_{\theta}(y \mid x)
\,\middle\|\,
\pi^{*}(y \mid x)
\right)
\right].
\label{eq:pre_opd_loss}
\end{equation}

Because the KL-divergence is computed at each token, OPD provides token-level supervision throughout the response.
Moreover, the supervision is applied to trajectories sampled from the current policy.
Note that the KL divergence is computed only for the sampled tokens. Therefore, similar to on-policy RL, OPD reduces the risk of the model deviating from its current generation distribution.
It adjusts the relative probabilities of candidate tokens within the model’s current distribution.

Given this connection between OPD and RL, the OPD objective can also be conveniently expressed using a token-level reward.
For a sampled token $y_t$, the OPD reward is defined as
\begin{equation}
R_t
=
\log \pi^{*}(y_t \mid x, y_{<t})
-
\log \pi_{\theta}(y_t \mid x, y_{<t}).
\label{eq:pre_opd_reward}
\end{equation}

Here, $(x, y_{<t})$ denotes the question together with the previously generated context, and $\pi(y_t \mid x, y_{<t})$ denotes the next-token probability assigned to the sampled token $y_t$.
Accordingly, OPD assigns positive rewards to tokens that are more strongly preferred by the teacher than by the student, while assigning negative rewards to tokens to which the teacher assigns relatively lower probability.

Using this token-level reward, OPD updates the policy model $\pi_\theta$ through the following objective:
\begin{equation}
\mathbb{E}_{y \sim \pi_{\theta}(\cdot \mid x)}
\left[
\sum_{t=1}^{T}
R_t
\nabla_{\theta}
\log \pi_{\theta}(y_t \mid x, y_{<t})
\right].
\label{eq:pre_opd_rl}
\end{equation}

\subsection{OPD$^2$}

On-Policy Delta Distillation (OPD$^2$) introduces a \textit{delta signal} as the reward function $R_t$ for OPD training. 
In general, LLMs are first trained with next-token prediction on large-scale pre-training data, resulting in a base model.
The base model is then post-trained with SFT and RL to acquire complex reasoning capabilities.
OPD$^2$ focuses on this post-training process and defines the difference between the teacher and its corresponding base model as the \textit{delta signal}.
This signal is designed to isolate the capabilities acquired during post-training, particularly reasoning ability, while reducing the influence of general preferences and stylistic patterns already present in the base model.

Denoting the teacher’s base model by $\pi^*_{\mathrm{base}}$, the OPD$^2$ reward is defined as
\begin{equation}
\log \pi^{*}(y_t \mid x, y_{<t})
-
\log \pi^{*}_{base}(y_t \mid x, y_{<t}).
\label{eq:delta_reward}
\end{equation}
Compared with Eq.~\ref{eq:pre_opd_reward}, OPD$^2$ simply replaces the student model $\pi_\theta$ in the second term with the teacher’s base model $\pi^*_{\mathrm{base}}$. 
Although this modification is simple, it brings substantial improvements.

To resolve the convergence-point issue of the \textit{delta signal}, OPD$^2$ introduces two additional components: reward centering to obtain the advantage and a conditioning criterion based on the original OPD signal. We incorporate both components in all our experiments, but omit their detailed formulations due to space constraints. We refer readers to \cite{heo2026opd2} for further details.

\begin{figure*}[t]
    \centering
    \includegraphics[width=\linewidth]{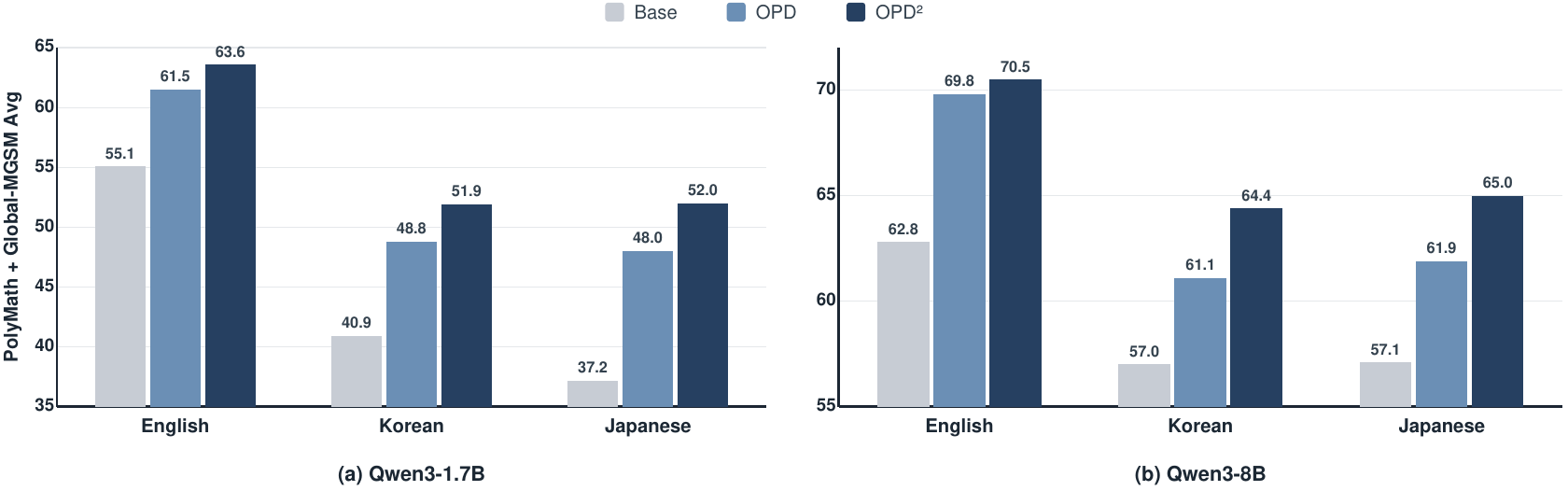}
    \caption{
        \textbf{Performance comparison for OPD and OPD$^2$.} The graph shows the mathematical reasoning benchmark results for PolyMath and Global-MGSM in three languages. OPD and OPD$^2$ substantially improve the performance for all languages.
    }
    \label{fig:opd_vs_opd2}
\end{figure*}

\section{Experiments}

We evaluate the improvements in mathematical reasoning achieved through OPD and OPD$^2$ training. We use instruction-tuned models and investigate whether OPD remains effective in multilingual settings. The complete results are provided in Appendix~\ref{sec:appendix}. In the main paper, we present a representative subset of the results to highlight the key performance trends across languages and benchmarks.

\subsection{Experiment settings}

We construct a multilingual mathematics training dataset containing 100K problems in total. It consists of English, Korean, and Japanese problems sampled at a ratio of 1:1:1, with no overlapping problems across languages. The Korean and Japanese problems are randomly sampled from the corresponding subsets of Nemotron-SFT-Multilingual-v2~\citep{nvidia2026nemotronmultilingualv2}, while the English problems are sampled from Nemotron-Math-v2~\citep{du2025nemotronmath}, which serves as the source dataset for the mathematical problems in Nemotron-SFT-Multilingual-v2. We additionally construct a separate English-only training set containing 100K English problems sampled from Nemotron-Math-v2. Since OPD does not require reference reasoning traces or answers, we retain only the question field from each example.

We conduct experiments using two models from the Qwen3 family~\citep{yang2025qwen3}, Qwen3-1.7B and Qwen3-8B, as student models. Qwen3-30B-A3B-2507 is used as the teacher model for distillation. We train and evaluate all models in both thinking and non-thinking modes. We evaluate the trained models on multilingual mathematical reasoning benchmarks: PolyMath~\citep{wang2026polymath} and Global-MGSM~\citep{shi2022MGSM,huang2025benchmax} in English, Korean, and Japanese; HRM8K~\citep{ko202haerae} in English and Korean; and MAWPS~\citep{horio2023MAWPS} in Japanese. 
We conduct all experiments using the official OPD$^2$ implementation,\footnote{\url{https://github.com/naver-ai/opd2}} and otherwise follow its training recipe~\citep{heo2026opd2}.
All models are trained for 100 optimization steps using the same hyperparameters and number of training examples across multilingual and English-only settings.

\begin{figure*}[t]
    \centering
    \includegraphics[width=\linewidth]{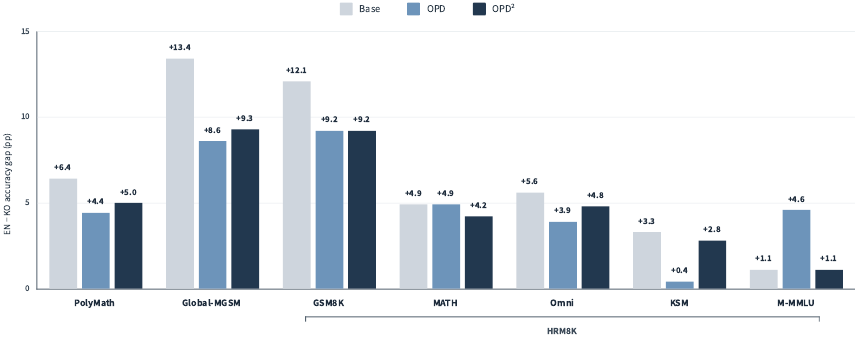}
    \caption{
        \textbf{English--Korean accuracy gap.} The graph shows the math performance difference for English and Korean in thinking mode. Multilingual OPD generally narrows the gap, although the effect varies across benchmarks.
    }
    \label{fig:en_ko_gap}
\end{figure*}

\subsection{Multilingual OPD}

Figure~\ref{fig:opd_vs_opd2} presents the average performance on PolyMath and Global-MGSM in non-thinking mode. Both OPD and OPD$^2$ improve mathematical reasoning across all model sizes and languages, demonstrating that on-policy distillation remains effective beyond English. For Qwen3-1.7B, OPD$^2$ improves the base-model average from 55.1 to 63.6 in English, from 40.9 to 51.9 in Korean, and from 37.2 to 52.0 in Japanese. The corresponding scores for Qwen3-8B increase from 62.8 to 70.5, from 57.0 to 64.4, and from 57.1 to 65.0, respectively. Thus, the improvements are observed not only for the smaller model, but also for the stronger Qwen3-8B model.

Moreover, OPD$^2$ consistently outperforms OPD, with particularly strong gains in Korean and Japanese. For Qwen3-1.7B, OPD$^2$ exceeds OPD by 3.1 and 4.0 points in Korean and Japanese, respectively; the corresponding gains for Qwen3-8B are 3.3 and 3.1 points. By comparison, the additional improvements over OPD in English are 2.1 points for Qwen3-1.7B and 0.7 points for Qwen3-8B. The advantage of the delta signal is therefore at least as pronounced in the non-English languages as in English. This suggests that OPD$^2$ does not merely imitate the teacher's general output distribution, but provides an effective post-training signal for transferring reasoning capabilities across languages and model scales.

\begin{figure*}[t]
    \centering
    \includegraphics[width=\linewidth]{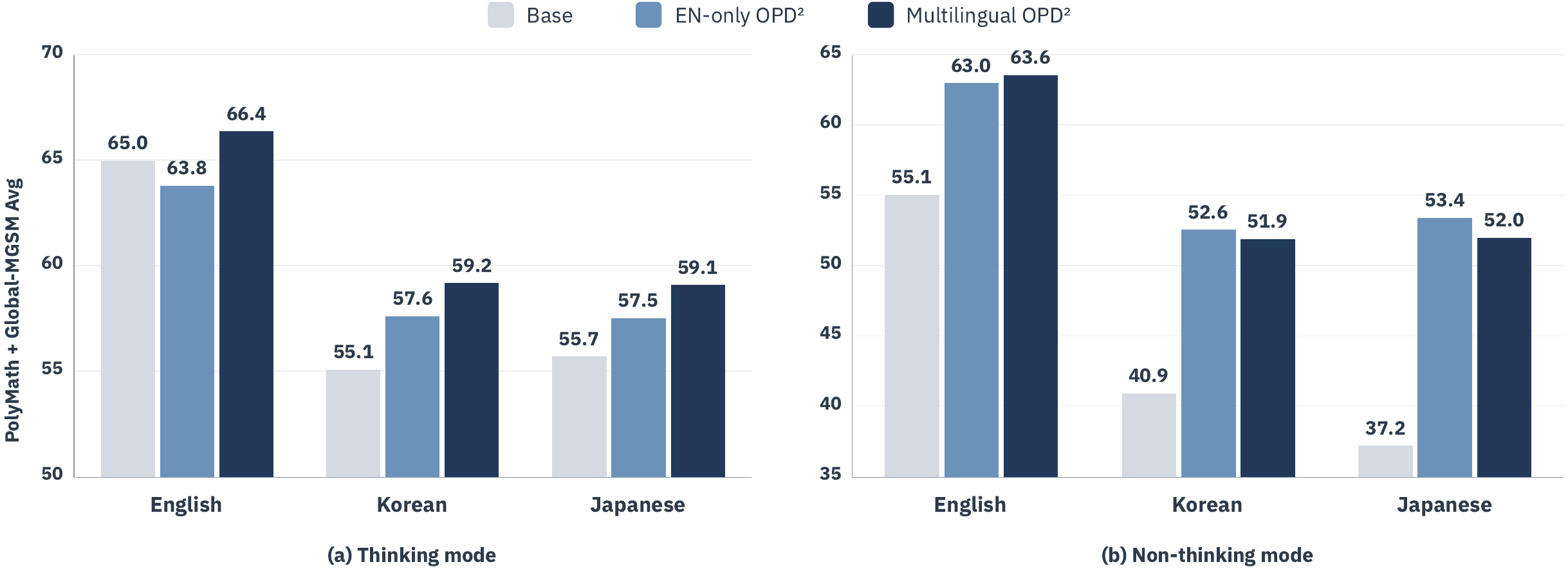}
    \caption{
        \textbf{English-only OPD$^2$ versus multilingual OPD$^2$.} We compare OPD$^2$ performance for the English-only dataset and the multilingual dataset. Surprisingly, English-only OPD$^2$ also improves the KO and JA performance, which is comparable to multilingual OPD$^2$ in non-thinking mode.
    }
    \label{fig:en_only}
\end{figure*}

\begin{table}[t]
\centering
\caption{
\textbf{Target-language response rates on PolyMath.}
We report the percentage of outputs generated in the question language.
For thinking models, the language rate is measured only after
\texttt{</think>} because models primarily use English for the think part. For non-thinking models, it is measured over the
entire response.
}
\label{tab:response_language}

\small
\renewcommand{\arraystretch}{1.12}
\setlength{\tabcolsep}{6pt} 

\begin{minipage}[t]{0.495\textwidth}
\centering
\textbf{(a) Thinking mode}\\[4pt]

\resizebox{\linewidth}{!}{
\begin{tabular}{lrrrr}
\toprule
& \multicolumn{2}{c}{OPD}
& \multicolumn{2}{c}{OPD$^2$} \\
\cmidrule(lr){2-3}
\cmidrule(lr){4-5}
Training data
& KO & JA
& KO & JA \\
\midrule
English-only
& 83.1\% & 36.9\%
& 36.1\% & 30.8\% \\
Multilingual
& 72.9\% & 70.1\%
& 97.6\% & 95.2\% \\
\bottomrule
\end{tabular}
}
\end{minipage}
\hfill
\begin{minipage}[t]{0.495\textwidth}
\centering
\textbf{(b) Non-thinking mode}\\[4pt]

\resizebox{\linewidth}{!}{
\begin{tabular}{lrrrr}
\toprule
& \multicolumn{2}{c}{OPD}
& \multicolumn{2}{c}{OPD$^2$} \\
\cmidrule(lr){2-3}
\cmidrule(lr){4-5}
Training data
& KO & JA
& KO & JA \\
\midrule
English-only
& 83.6\% & 40.5\%
& 48.3\% & 29.6\% \\
Multilingual
& 99.7\% & 99.8\%
& 90.5\% & 90.9\% \\
\bottomrule
\end{tabular}
}
\end{minipage}

\end{table}

\subsection{English--Korean Performance Gap}

Figure~\ref{fig:en_ko_gap} compares English and Korean accuracy for Qwen3-1.7B in thinking mode. The base model consistently favors English, with particularly large gaps of 13.4 points on Global-MGSM and 12.1 points on HRM8K-GSM8K. OPD generally narrows this disparity: for example, the gaps decrease from 6.4 to 4.4 points on PolyMath and from 13.4 to 8.6 points on Global-MGSM. The gap on KSM is also substantially reduced from 3.3 to 0.4 points. However, the effect is not completely uniform across benchmarks, as the gap on M-MMLU increases from 1.1 to 4.6 points after OPD training.

OPD$^2$ shows a more consistent reduction in the language gap. It decreases the gaps on PolyMath and Global-MGSM to 5.0 and 9.3 points, respectively, and reduces the HRM8K-GSM8K gap from 12.1 to 9.2 points. It also narrows the gaps on MATH, Omni, and KSM, while leaving the already small M-MMLU gap unchanged at 1.1 points. Overall, OPD$^2$ reduces the English--Korean gap on six of the seven reported benchmarks and preserves it on the remaining benchmark. These results indicate that multilingual post-training often benefits Korean reasoning more strongly than English reasoning, thereby reducing cross-lingual disparity without sacrificing the strong English performance.

\subsection{Comparison with English-Only OPD$^2$}
\label{sec:eng_only}

Figure~\ref{fig:en_only} compares multilingual and English-only OPD$^2$. Surprisingly, English-only OPD$^2$ also improves mathematical reasoning performance in Korean and Japanese. In non-thinking mode, it improves their average scores from 40.9 to 52.6 and from 37.2 to 53.4, respectively. These results are comparable to the multilingual OPD$^2$ scores of 51.9 in Korean and 52.0 in Japanese. In thinking mode, English-only OPD$^2$ similarly improves the Korean and Japanese averages to 57.6 and 57.5, although multilingual training achieves somewhat higher scores of 59.2 and 59.1. This indicates that reasoning capabilities learned through English-only distillation can transfer to non-English inputs, even without directly observing Korean or Japanese training questions.

However, benchmark accuracy alone does not fully reflect the language used by the model when producing its solution. We therefore measure the target-language response rate, defined as the proportion of outputs whose generated text is identified as matching the language of the input question. The results are summarized in Table~\ref{tab:response_language}. For thinking models, we measure the response language only after \texttt{</think>}, because the intermediate reasoning traces are primarily generated in English. For non-thinking models, the language is measured over the entire response.

As shown in Table~\ref{tab:response_language}, replacing multilingual training with English-only training substantially reduces the target-language response rate of OPD$^2$. In non-thinking mode, the Korean response rate decreases from 90.5\% to 48.3\%, while the Japanese response rate decreases from 90.9\% to 29.6\%. The same tendency is observed for final answers in thinking mode: the Korean rate drops from 97.6\% to 36.1\%, and the Japanese rate drops from 95.2\% to 30.8\%. In other words, the English-only models can obtain competitive Korean and Japanese benchmark accuracy while frequently responding in English rather than in the language of the question.

Table~\ref{tab:response_language} also shows that this behavior is not determined solely by whether the model operates in thinking or non-thinking mode. Across both modes, multilingual OPD$^2$ preserves substantially higher Korean and Japanese response rates than its English-only counterpart. These results distinguish the transfer of mathematical reasoning ability from the preservation of target-language generation. English-only OPD$^2$ can transfer reasoning performance across languages, but multilingual training remains important for ensuring that the acquired capability is expressed in the requested language.



\section{Conclusion}

We study OPD and OPD$^2$ for mathematical reasoning in English, Korean, and Japanese. Both methods improve multilingual performance across model sizes and generation modes, while OPD$^2$ consistently outperforms the original OPD and generally reduces the English--Korean performance gap. English-only training can also transfer reasoning accuracy to Korean and Japanese, showing that reasoning capabilities learned from English supervision can generalize across languages. However, these gains often come with a shift in the generated response toward English, even when the input is given in Korean or Japanese. These results demonstrate the effectiveness of OPD$^2$ for multilingual post-training, distinguish cross-lingual reasoning transfer from target-language generation, and highlight the importance of evaluating both benchmark accuracy and response language in multilingual reasoning models.

{\small
\bibliographystyle{unsrt}
\bibliography{custom}}

\newpage
\appendix

\section{Full Experimental Results}
\label{sec:appendix}

We report the complete benchmark results for Qwen3-1.7B and Qwen3-8B in both thinking and non-thinking modes. The results are organized by the language of the evaluation benchmark. For English and Korean, we report PolyMath, Global-MGSM, and the five subsets of HRM8K: GSM8K, MATH, Omni, KSM, and M-MMLU. For Japanese, we report PolyMath, Global-MGSM, and MAWPS. The average score is computed over all benchmarks presented in each table.

Unless otherwise specified, OPD and OPD$^2$ are trained on the balanced multilingual dataset containing English, Korean, and Japanese questions at a ratio of 1:1:1. We additionally report an English-only training ablation for Qwen3-1.7B. The base rows correspond to the original instruction-tuned models without additional OPD post-training.

\subsection{Qwen3-1.7B with Multilingual Training}

Tables~\ref{tab:ml-q17b-en}--\ref{tab:ml-q17b-ja} present the full results for Qwen3-1.7B trained on the multilingual dataset. Both OPD and OPD$^2$ substantially improve the non-thinking performance across all three languages. OPD$^2$ achieves the highest average score in English, Korean, and Japanese, improving the corresponding base model averages from 47.6 to 65.7, from 37.4 to 56.5, and from 55.5 to 65.4, respectively.

The improvements in thinking mode are smaller because the base model already exhibits strong reasoning performance. Nevertheless, OPD$^2$ consistently improves the overall average, reaching 73.4 in English, 68.2 in Korean, and 71.2 in Japanese. In comparison, the original OPD obtains 71.4, 66.2, and 70.5, respectively. These results further support the consistent advantage of OPD$^2$ over the original OPD.


\subsection{Qwen3-1.7B with English-Only Training}

Tables~\ref{tab:mlen-q17b-en}--\ref{tab:mlen-q17b-ja} report the results of Qwen3-1.7B trained using only English questions. In non-thinking mode, English-only training transfers effectively to Korean and Japanese benchmarks. In particular, English-only OPD$^2$ improves the average score from 37.4 to 59.3 in Korean and from 55.5 to 66.5 in Japanese. These scores are comparable to, and in some cases higher than, those obtained using multilingual training.

The behavior differs in thinking mode. English-only OPD substantially degrades the average Korean score from 63.7 to 48.5, whereas OPD$^2$ retains and slightly improves the base performance, reaching 65.1. On the English benchmarks, neither English-only OPD nor OPD$^2$ improves the overall thinking-mode average over the base model. These results indicate that the effect of English-only post-training depends on both the distillation objective and the generation mode.

As discussed in Section~\ref{sec:eng_only}, strong benchmark accuracy under English-only training does not necessarily imply that the model reasons or responds in the target language. Our response-language analysis shows that English-only models frequently generate English text for Korean and Japanese PolyMath inputs, particularly after OPD$^2$ training.


\subsection{Qwen3-8B with Multilingual Training}

Tables~\ref{tab:ml-q8b-en}--\ref{tab:ml-q8b-ja} provide the complete results for Qwen3-8B. In non-thinking mode, OPD$^2$ improves the average score from 58.7 to 75.3 in English, from 55.4 to 72.2 in Korean, and from 70.2 to 75.2 in Japanese. It also consistently outperforms the original OPD in all three languages.

For thinking mode, the original OPD occasionally reduces performance relative to the strong base model. For example, its average score decreases from 80.9 to 79.9 in English and from 78.0 to 75.7 in Korean. In contrast, OPD$^2$ improves the corresponding averages to 83.2 and 80.4. OPD$^2$ also obtains the best Japanese average of 78.9. These results suggest that OPD$^2$ provides a more reliable post-training signal than the original OPD, especially when the initial model already has strong reasoning capabilities.


%

\begin{table*}[t]
\centering
\caption{English benchmarks for Qwen3-1.7B trained with multilingual OPD.}
\label{tab:ml-q17b-en}
\small
\resizebox{\textwidth}{!}{%
\begin{tabular}{llccccccc c}
\toprule
& & & & \multicolumn{5}{c}{HRM8K-EN} & \\
\cmidrule(lr){5-9}
Mode & Model & PolyMath-EN & GMGSM-EN & GSM8K & MATH & Omni & KSM & M-MMLU & Avg \\
\midrule
\multirow{3}{*}{\begin{tabular}[c]{@{}l@{}}Non-\\Thinking\end{tabular}}
 & Qwen3-1.7B      & 25.5 & 84.8 & 80.1 & 74.4 & 29.1 & 21.6 & 17.9 & 47.6 \\
 & \quad + OPD     & 32.2 & 90.7 & 87.3 & 88.0 & 44.0 & 42.7 & 47.0 & 61.7 \\
 & \quad + OPD$^2$ & \textbf{34.3} & \textbf{92.8} & \textbf{89.0} & \textbf{91.1} & \textbf{48.7} & \textbf{48.8} & \textbf{55.3} & \textbf{65.7} \\
\midrule
\multirow{3}{*}{Thinking}
 & Qwen3-1.7B      & 36.3 & 93.7 & 90.5 & 92.6 & 52.8 & 55.3 & 71.6 & 70.4 \\
 & \quad + OPD     & 36.3 & 94.0 & 90.3 & 93.7 & 52.7 & 53.7 & \textbf{78.8} & 71.4 \\
 & \quad + OPD$^2$ & \textbf{37.7} & \textbf{95.1} & \textbf{91.2} & \textbf{94.9} & \textbf{55.3} & \textbf{61.2} & 78.1 & \textbf{73.4} \\
\bottomrule
\end{tabular}%
}
\end{table*}

\begin{table*}[t]
\centering
\caption{Korean benchmarks for Qwen3-1.7B trained with multilingual OPD.}
\label{tab:ml-q17b-ko}
\small
\resizebox{\textwidth}{!}{%
\begin{tabular}{llccccccc c}
\toprule
& & & & \multicolumn{5}{c}{HRM8K-KO} & \\
\cmidrule(lr){5-9}
Mode & Model & PolyMath-KO & GMGSM-KO & GSM8K & MATH & Omni & KSM & M-MMLU & Avg \\
\midrule
\multirow{3}{*}{\begin{tabular}[c]{@{}l@{}}Non-\\Thinking\end{tabular}}
 & Qwen3-1.7B      & 18.0 & 63.7 & 61.9 & 57.3 & 19.5 & 13.9 & 27.6 & 37.4 \\
 & \quad + OPD     & 23.4 & 74.2 & 71.1 & 71.9 & 28.2 & 26.8 & 45.5 & 48.7 \\
 & \quad + OPD$^2$ & \textbf{26.5} & \textbf{77.2} & \textbf{73.9} & \textbf{81.0} & \textbf{37.9} & \textbf{36.5} & \textbf{62.4} & \textbf{56.5} \\
\midrule
\multirow{3}{*}{Thinking}
 & Qwen3-1.7B      & 29.9 & 80.3 & 78.4 & 87.8 & 47.2 & 52.0 & 70.5 & 63.7 \\
 & \quad + OPD     & 31.9 & 85.4 & 81.1 & 88.8 & 48.8 & 53.4 & 74.2 & 66.2 \\
 & \quad + OPD$^2$ & \textbf{32.7} & \textbf{85.8} & \textbf{82.0} & \textbf{90.7} & \textbf{50.5} & \textbf{58.4} & \textbf{77.0} & \textbf{68.2} \\
\bottomrule
\end{tabular}%
}
\end{table*}

\begin{table*}[t]
\centering
\caption{Japanese benchmarks for Qwen3-1.7B trained with multilingual OPD.}
\label{tab:ml-q17b-ja}
\small
\resizebox{\textwidth}{!}{%
\begin{tabular}{lcccc cccc}
\toprule
& \multicolumn{4}{c}{Non-Thinking} & \multicolumn{4}{c}{Thinking} \\
\cmidrule(lr){2-5} \cmidrule(lr){6-9}
Model & PolyMath-JA & GMGSM-JA & MAWPS & Avg & PolyMath-JA & GMGSM-JA & MAWPS & Avg \\
\midrule
Qwen3-1.7B      & 16.3 & 58.2 & 92.2 & 55.5 & 30.5 & 80.9 & 94.2 & 68.5 \\
\quad + OPD     & 22.7 & 73.3 & \textbf{92.8} & 62.9 & 31.4 & \textbf{85.1} & 95.0 & 70.5 \\
\quad + OPD$^2$ & \textbf{25.6} & \textbf{78.3} & 92.4 & \textbf{65.4} & \textbf{33.2} & \textbf{85.1} & \textbf{95.3} & \textbf{71.2} \\
\bottomrule
\end{tabular}%
}
\end{table*}


\begin{table*}[t]
\centering
\caption{English benchmarks for Qwen3-1.7B trained with English-only OPD.}
\label{tab:mlen-q17b-en}
\small
\resizebox{\textwidth}{!}{%
\begin{tabular}{llccccccc c}
\toprule
& & & & \multicolumn{5}{c}{HRM8K-EN} & \\
\cmidrule(lr){5-9}
Mode & Model & PolyMath-EN & GMGSM-EN & GSM8K & MATH & Omni & KSM & M-MMLU & Avg \\
\midrule
\multirow{3}{*}{\begin{tabular}[c]{@{}l@{}}Non-\\Thinking\end{tabular}}
 & Qwen3-1.7B      & 25.5 & 84.8 & 80.1 & 74.4 & 29.1 & 21.6 & 17.9 & 47.6 \\
 & \quad + OPD     & 32.5 & 91.4 & 86.4 & 87.9 & 44.2 & 40.7 & 60.3 & 63.3 \\
 & \quad + OPD$^2$ & \textbf{34.3} & \textbf{91.7} & \textbf{88.7} & \textbf{91.5} & \textbf{49.5} & \textbf{51.1} & \textbf{61.8} & \textbf{66.9} \\
\midrule
\multirow{3}{*}{Thinking}
 & Qwen3-1.7B      & \textbf{36.3} & \textbf{93.7} & \textbf{90.5} & 92.6 & \textbf{52.8} & \textbf{55.3} & 71.6 & \textbf{70.4} \\
 & \quad + OPD     & 34.4 & 92.9 & 90.0 & 92.2 & 49.4 & 49.9 & \textbf{77.2} & 69.4 \\
 & \quad + OPD$^2$ & 34.3 & 93.3 & 89.6 & \textbf{93.0} & 48.5 & 50.1 & 76.1 & 69.3 \\
\bottomrule
\end{tabular}%
}
\end{table*}

\begin{table*}[t]
\centering
\caption{Korean benchmarks for Qwen3-1.7B trained with English-only OPD.}
\label{tab:mlen-q17b-ko}
\small
\resizebox{\textwidth}{!}{%
\begin{tabular}{llccccccc c}
\toprule
& & & & \multicolumn{5}{c}{HRM8K-KO} & \\
\cmidrule(lr){5-9}
Mode & Model & PolyMath-KO & GMGSM-KO & GSM8K & MATH & Omni & KSM & M-MMLU & Avg \\
\midrule
\multirow{3}{*}{\begin{tabular}[c]{@{}l@{}}Non-\\Thinking\end{tabular}}
 & Qwen3-1.7B      & 18.0 & 63.7 & 61.9 & 57.3 & 19.5 & 13.9 & 27.6 & 37.4 \\
 & \quad + OPD     & 23.2 & 72.2 & 70.6 & 72.6 & 29.6 & 28.1 & 54.8 & 50.1 \\
 & \quad + OPD$^2$ & \textbf{26.9} & \textbf{78.4} & \textbf{75.1} & \textbf{81.6} & \textbf{40.4} & \textbf{43.4} & \textbf{69.1} & \textbf{59.3} \\
\midrule
\multirow{3}{*}{Thinking}
 & Qwen3-1.7B      & 29.9 & 80.3 & 78.4 & 87.8 & \textbf{47.2} & \textbf{52.0} & 70.5 & 63.7 \\
 & \quad + OPD     & 22.8 & 72.3 & 69.0 & 67.7 & 26.9 & 24.4 & 56.5 & 48.5 \\
 & \quad + OPD$^2$ & \textbf{30.3} & \textbf{84.8} & \textbf{80.6} & \textbf{88.8} & 45.8 & 49.0 & \textbf{76.5} & \textbf{65.1} \\
\bottomrule
\end{tabular}%
}
\end{table*}

\begin{table*}[t]
\centering
\caption{Japanese benchmarks for Qwen3-1.7B trained with English-only OPD.}
\label{tab:mlen-q17b-ja}
\small
\resizebox{\textwidth}{!}{%
\begin{tabular}{lcccc cccc}
\toprule
& \multicolumn{4}{c}{Non-Thinking} & \multicolumn{4}{c}{Thinking} \\
\cmidrule(lr){2-5} \cmidrule(lr){6-9}
Model & PolyMath-JA & GMGSM-JA & MAWPS & Avg & PolyMath-JA & GMGSM-JA & MAWPS & Avg \\
\midrule
Qwen3-1.7B      & 16.3 & 58.2 & 92.2 & 55.5 & \textbf{30.5} & 80.9 & 94.2 & 68.5 \\
\quad + OPD     & 24.6 & 72.2 & 91.9 & 62.9 & 29.8 & 83.5 & 94.5 & 69.3 \\
\quad + OPD$^2$ & \textbf{27.9} & \textbf{79.0} & \textbf{92.7} & \textbf{66.5} & 30.1 & \textbf{84.9} & \textbf{94.7} & \textbf{69.9} \\
\bottomrule
\end{tabular}%
}
\end{table*}

%
%

\begin{table*}[t]
\centering
\caption{English benchmarks for Qwen3-8B trained with multilingual OPD.}
\label{tab:ml-q8b-en}
\small
\resizebox{\textwidth}{!}{%
\begin{tabular}{llccccccc c}
\toprule
& & & & \multicolumn{5}{c}{HRM8K-EN} & \\
\cmidrule(lr){5-9}
Mode & Model & PolyMath-EN & GMGSM-EN & GSM8K & MATH & Omni & KSM & M-MMLU & Avg \\
\midrule
\multirow{3}{*}{\begin{tabular}[c]{@{}l@{}}Non-\\Thinking\end{tabular}}
 & Qwen3-8B        & 31.6 & 93.9 & 88.1 & 87.0 & 42.6 & 38.7 & 28.8 & 58.7 \\
 & \quad + OPD     & 41.2 & \textbf{98.4} & 93.3 & 95.2 & 63.9 & 67.2 & \textbf{58.9} & 74.0 \\
 & \quad + OPD$^2$ & \textbf{42.9} & 98.0 & \textbf{93.3} & \textbf{96.4} & \textbf{68.4} & \textbf{73.4} & 54.5 & \textbf{75.3} \\
\midrule
\multirow{3}{*}{Thinking}
 & Qwen3-8B        & 43.5 & \textbf{98.8} & \textbf{95.7} & 97.5 & 69.6 & 76.5 & 84.9 & 80.9 \\
 & \quad + OPD     & 42.3 & 98.4 & 95.4 & 97.7 & 68.8 & 75.2 & 81.4 & 79.9 \\
 & \quad + OPD$^2$ & \textbf{44.5} & 98.3 & \textbf{95.7} & \textbf{98.2} & \textbf{72.8} & \textbf{79.1} & \textbf{93.9} & \textbf{83.2} \\
\bottomrule
\end{tabular}%
}
\end{table*}

\begin{table*}[t]
\centering
\caption{Korean benchmarks for Qwen3-8B trained with multilingual OPD.}
\label{tab:ml-q8b-ko}
\small
\resizebox{\textwidth}{!}{%
\begin{tabular}{llccccccc c}
\toprule
& & & & \multicolumn{5}{c}{HRM8K-KO} & \\
\cmidrule(lr){5-9}
Mode & Model & PolyMath-KO & GMGSM-KO & GSM8K & MATH & Omni & KSM & M-MMLU & Avg \\
\midrule
\multirow{3}{*}{\begin{tabular}[c]{@{}l@{}}Non-\\Thinking\end{tabular}}
 & Qwen3-8B        & 27.7 & 86.2 & 85.4 & 80.1 & 33.7 & 31.1 & 43.4 & 55.4 \\
 & \quad + OPD     & 33.3 & 88.9 & 87.0 & 89.0 & 47.6 & 53.2 & \textbf{70.4} & 67.1 \\
 & \quad + OPD$^2$ & \textbf{38.3} & \textbf{90.4} & \textbf{88.2} & \textbf{92.5} & \textbf{61.4} & \textbf{70.8} & 63.9 & \textbf{72.2} \\
\midrule
\multirow{3}{*}{Thinking}
 & Qwen3-8B        & 41.2 & 92.8 & 91.6 & 95.5 & 67.2 & 75.6 & 82.1 & 78.0 \\
 & \quad + OPD     & 39.4 & 92.1 & 91.5 & 94.9 & 64.6 & 74.8 & 72.7 & 75.7 \\
 & \quad + OPD$^2$ & \textbf{41.7} & \textbf{93.6} & \textbf{91.8} & \textbf{95.9} & \textbf{70.7} & \textbf{80.0} & \textbf{89.1} & \textbf{80.4} \\
\bottomrule
\end{tabular}%
}
\end{table*}

\begin{table*}[t]
\centering
\caption{Japanese benchmarks for Qwen3-8B trained with multilingual OPD.}
\label{tab:ml-q8b-ja}
\small
\resizebox{\textwidth}{!}{%
\begin{tabular}{lcccc cccc}
\toprule
& \multicolumn{4}{c}{Non-Thinking} & \multicolumn{4}{c}{Thinking} \\
\cmidrule(lr){2-5} \cmidrule(lr){6-9}
Model & PolyMath-JA & GMGSM-JA & MAWPS & Avg & PolyMath-JA & GMGSM-JA & MAWPS & Avg \\
\midrule
Qwen3-8B        & 26.9 & 87.3 & \textbf{96.4} & 70.2 & 39.9 & 96.5 & \textbf{96.6} & 77.7 \\
\quad + OPD     & 33.5 & 90.2 & 96.4 & 73.4 & 40.5 & 96.4 & 96.1 & 77.6 \\
\quad + OPD$^2$ & \textbf{37.9} & \textbf{92.0} & 95.8 & \textbf{75.2} & \textbf{42.9} & \textbf{97.5} & 96.4 & \textbf{78.9} \\
\bottomrule
\end{tabular}%
}
\end{table*}

\end{document}